\documentclass{mylatex2e}

\usepackage{cite}
\usepackage{graphicx}
\usepackage{amsfonts}
\usepackage{stmaryrd}
\usepackage{mathdots}
\usepackage{epsf}
\usepackage{theorem}
\usepackage{amsmath}
\usepackage{amssymb}
\usepackage{textcomp}
\usepackage{bm}
\usepackage{pdflscape}

\theoremstyle{plain}

\begin{document}

\renewcommand{\baselinestretch}{1.4}
\small\normalsize

\runauthor{Z.-H. Zhou et al.}
\begin{frontmatter}

\title{Deep Forest}\vspace{-2mm}

\renewcommand{\thefootnote}{\fnsymbol{footnote}}

\large{Zhi-Hua Zhou, Ji Feng}\vspace{+5mm}

{\small \textit{National Key Laboratory for Novel Software Technology,\\Nanjing
University, Nanjing 210023, China
\\ \{zhouzh, fengj\}@lamda.nju.edu.cn
}}\vspace{+5mm}

\renewcommand{\thefootnote}{\arabic{footnote}}
\setcounter{footnote}{0}

\begin{abstract}
Current deep learning models are mostly build upon neural networks, i.e., multiple layers of parameterized differentiable nonlinear modules that can be trained by backpropagation. In this paper, we explore the possibility of building deep models based on non-differentiable modules. We conjecture that the mystery behind the success of deep neural networks owes much to three characteristics, i.e., layer-by-layer processing, in-model feature transformation and sufficient model complexity. We propose the gcForest approach, which generates \textit{deep forest} holding these characteristics. This is a decision tree ensemble approach, with much less hyper-parameters than deep neural networks, and its model complexity can be automatically determined in a data-dependent way. Experiments show that its performance is quite robust to hyper-parameter settings, such that in most cases, even across different data from different domains, it is able to get excellent performance by using the same default setting. This study opens the door of deep learning based on non-differentiable modules, and exhibits the possibility of constructing deep models without using backpropagation.
\end{abstract}


\end{frontmatter}


\section{Introduction}\vspace{-4mm}

Deep learning \cite{Goodfellow:Bengio2016} has become a hotwave in various domains. While, what is deep learning? Answers from the crowd are very likely to be that ``deep learning is a subfield of machine learning that uses deep neural networks'' \cite{SIAMnews2017}. Actually, the great success of deep neural networks (DNNs) in tasks involving visual and speech information \cite{AlexNet2012,Hinton:Deng2012} led to the rise of deep learning, and almost all current deep learning applications are built upon neural network models, or more technically, multiple layers of parameterized differentiable nonlinear modules that can be trained by backpropagation.

Though deep neural networks are powerful, they have many deficiencies. First, DNNs are with too many hyper-parameters, and the learning performance depends seriously on careful parameter tuning. Indeed, even when several authors all use convolutional neural networks \cite{LeCun:Bottou1998,AlexNet2012,Simonyan:Zisserman2014}, they are actually using different learning models due to the many different options such as the convolutional layer structures. This fact makes not only the training of DNNs very tricky, almost like an art rather than science/engineering, but also theoretical analysis of DNNs extremely difficult because of too many interfering factors with almost infinite configurational combinations. Second, it is well known that the training of DNNs requires a huge amount of training data, and thus, DNNs can hardly be applied to tasks where there are only small-scale training data, sometimes even fail with mid-scale training data. Note that even in the big data era, many real tasks still lack sufficient amount of \textit{labeled} data due to the high cost of labeling, leading to inferior performance of DNNs in these tasks. Moreover, it is well known that neural networks are black-box models whose decision processes are hard to understand, and the learning behaviors are very difficult for theoretical analysis. Furthermore, before training the neural network architecture has to be determined, and thus, the model complexity is determined in advance. We conjecture that deep models are usually overly complicated than what are really needed, as verified by the observation that recently there are many reports about DNNs performance improvement by adding shortcut connection \cite{Srivastava:Greff:Schmidhuber2015,he2016deep}, pruning \cite{Han:Pool:Tran2015,Luo:Wu:Lin2017}, binarization \cite{Courbariaux:Bengio:David2015,Rastegari:Ordonez:Redmon2016}, etc., because these operations simplify the original networks and actually decrease model complexity. It might be better if the model complexity can be determined automatically in a data-dependent way. It is also noteworthy that although DNNs have been well developed, there are still many tasks on which DNNs are not superior, sometimes even inadequate; for example, Random Forest \cite{Breiman2001} or XGBoost \cite{Chen2016xgboost} are still winners on many Kaggle competition tasks.

We believe that in order to tackle complicated learning tasks, learning models are likely have to go deep. Current deep models, however, are always build upon neural networks. As discussed above, there are good reasons to explore non-NN style deep models, or in other words, to consider whether deep learning can be realized with other modules, as they have their own advantages and may exhibit great potentials if being able to go deep. In particular, considering that neural networks are multiple layers of parameterized \textit{differentiable} nonlinear modules, whereas not all properties in the world are differentiable or best modelled as differentiable, in this paper we attempt to address this fundamental question:
\begin{center}
``\textit{Can deep learning be realized with non-differentiable modules}?''
\end{center}
The result may help understand many important issues such as (1) deep models ?= DNNs (or, deep models can only be constructed with differentiable modules); (2) Is it possible to train deep models without backpropagation? (backpropagation requires differentiability); (3) Is it possible to enable deep models win tasks on which now other models such as random forest or XGBoost are better? Actually, the machine learning community have developed lots of learning modules, whereas many of them are non-differentiable; understanding whether it is possible to construct deep models based on non-differentiable modules will give light on the issue that whether these modules can be exploited in deep learning.

In this paper, we extend our preliminary study \cite{Zhou:Feng2018} which proposes the gcForest\footnote{Sounds like ``geek forest''.} (multi-Grained Cascade Forest) approach for constructing deep forest, a non-NN style deep model. This is a novel decision tree ensemble, with a cascade structure which enables representation learning by forests. Its representational learning ability can be further enhanced by multi-grained scanning, potentially enabling gcForest to be contextual or structural aware. The cascade levels can be automatically determined such that the model complexity can be determined in a data-dependent way rather than manually designed before training; this enables gcForest to work well even on small-scale data, and enables users to control training costs according to computational resource available. Moreover, the gcForest has much fewer hyper-parameters than DNNs. Even better news is that its performance is quite robust to hyper-parameter settings; our experiments show that in most cases, it is able to get excellent performance by using the default setting, even across different data from different domains.

The rest of this paper is organized as follows. Section~\ref{sec:inspiration} explains our design motivations by analyzing why deep learning works. Section~\ref{sec:approach} proposes our approach, followed by experiments reported in Section~\ref{sec:experiments}. Section~\ref{sec:related} discusses on some related work. Section~\ref{sec:future} raises some issues for future exploration, followed by concluding remarks in Section~\ref{sec:conclusion}.

\section{Inspiration}\label{sec:inspiration}\vspace{-4mm}

\subsection{Inspiration from DNNs}\vspace{-4mm}

\begin{figure}[!ht]
\centering
\includegraphics[width=.75\textwidth]{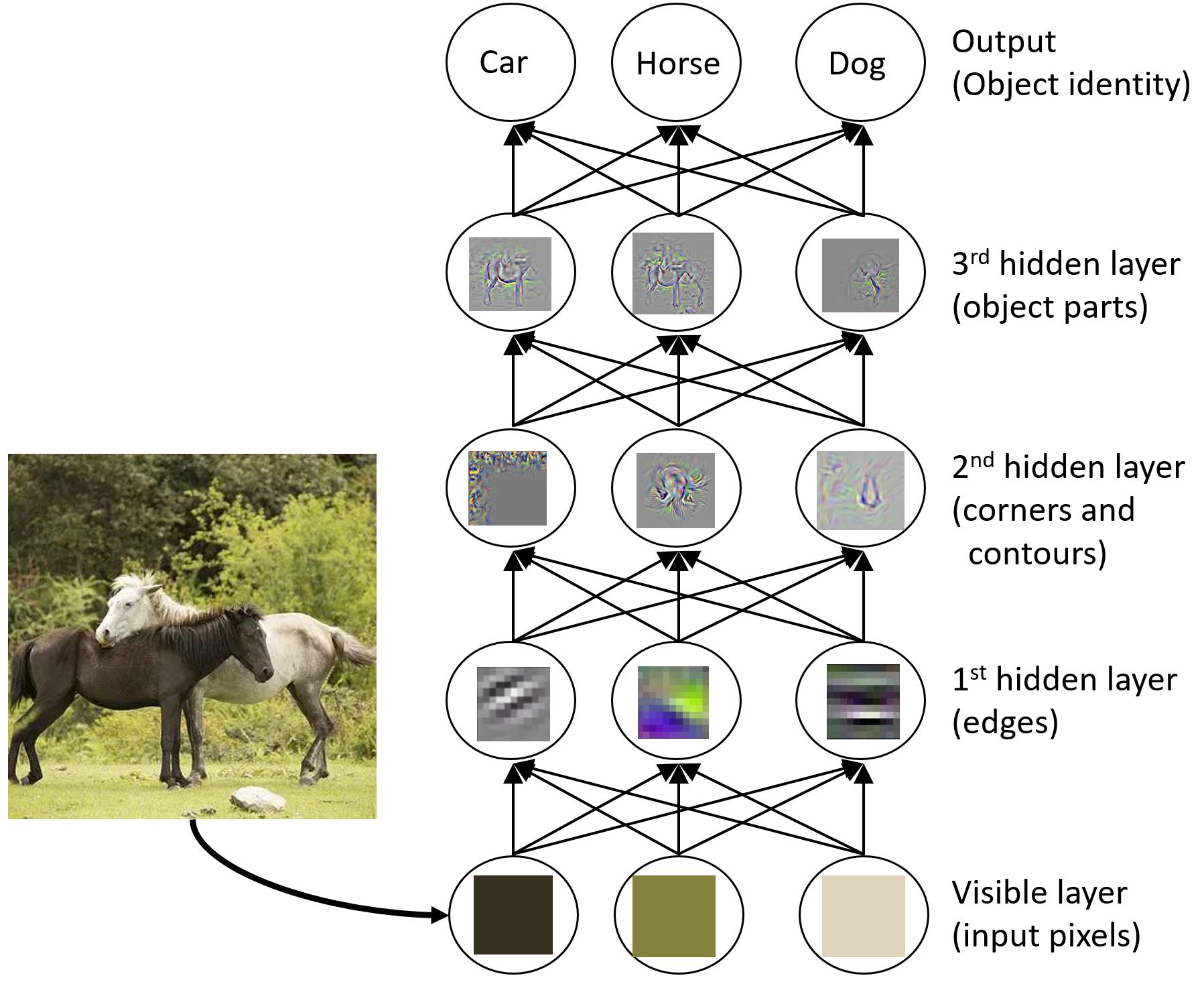}
\caption{Illustration of the layer-by-layer processing in deep neural networks: Features of higher levels of abstract emerge as the layer goes up from the bottom. Simulated from a figure in \cite{Goodfellow:Bengio2016}.}\label{fig:layer}\bigskip
\end{figure}

It is widely recognized that the \textit{representation learning} ability is crucial for the success of deep neural networks. While, what is crucial for representation learning in DNNs? We believe that the answer is \textit{layer-by-layer processing}. Figure~\ref{fig:layer} provides an illustration, where features of higher levels of abstract emerge as the layer goes up from the bottom.

Considering that if other issues fixed, large model complexity (or more accurately, model capacity) generally leads to strong learning ability, it sounds reasonable to attribute the successfulness of DNNs to the huge model complexity. This, however, cannot explain the fact that why shallow networks are not as successful as deep ones, as one can increase the complexity of shallow networks by adding nearly infinite number of hidden units. Thus, we believe that the model complexity itself cannot explain the success of DNNs. Instead, we conjecture that the layer-by-layer processing is one of the most important factors behind DNNs, because flat networks (e.g., single-hidden-layer networks), no matter how large their complexity can be, do not hold the characteristics of layer-by-layer processing. Although we do not have a rigorous justification yet, this conjecture offers an important inspiration for the design of gcForest.

One may question that there are learning models, e.g., decision trees and Boosting machines, which also conduct layer-by-layer processing, why they are not as successful as DNNs? We believe that the most important distinguishing factor is that, in contrast to DNNs where new features are generated as illustrated in Figure~\ref{fig:layer}, decision trees and Boosting machines always work on the original feature representation without creating new features during the learning process, or in other words, there is no in-model feature transformation. Moreover, in contrast to DNNs that can be endowed with arbitrarily high model complexity, decision trees and Boosting machines can only have limited model complexity. Although the model complexity itself does not necessarily explain the successfulness of DNNs, it is still important because large model capacity is needed for exploiting large training data.

Overall, we conjecture that behind the mystery of DNNs there are three crucial characteristics, i.e., layer-by-layer processing, in-model feature transformation, and sufficient model complexity. We will try to endow these characteristics to our non-NN style deep model.

\subsection{Inspiration from Ensemble Learning}\vspace{-5mm}

Ensemble learning \cite{Zhou2012} is a machine learning paradigm where multiple learners (e.g., classifiers) are trained and combined for a task. It is well known that an ensemble can usually achieve better generalization performance than single learners.

To construct a good ensemble, the individual learners should be \textit{accurate} and \textit{diverse}. Combining only accurate learners is often inferior to combining some accurate learners with some relatively weaker ones, because the complementarity is more important than pure accuracy. Actually, a beautiful equation has been theoretically derived from \textit{error-ambiguity decomposition} \cite{Krogh:Vedelsby1995}:
\begin{equation}\label{eq:ensemble}
E = \bar{E} - \bar{A} \ ,
\end{equation}
where $E$ denotes the error of an ensemble, $\bar{E}$ denotes the average error of individual classifiers in the ensemble, and $\bar{A}$ denotes the average \textit{ambiguity}, later called \textit{diversity}, among the individual classifiers. Eq.~\ref{eq:ensemble} reveals that, the more accurate and more diverse the individual classifiers, the better the ensemble.
This offers a general guidance for ensemble construction; however, it could not be taken as an objective function for optimization, because the \textit{ambiguity} term is mathematically defined in the derivation and cannot be operated directly \cite{Krogh:Vedelsby1995}. Later on, the ensemble community have designed lots of diversity measures, but none has been well-accepted as the \textit{right} definition for diversity \cite{Kuncheva:Whitaker2003,Didaci:Fumera:Roli2013}. Actually, ``\textit{what is diversity}?'' remains the holy grail problem in ensemble learning, and some recent effort can be found in \cite{Zhou:Li2010,Sun:Zhou2018}.

In practice, the basic strategy of diversity enhancement is to inject randomness based on some heuristics during the training process. Roughly speaking, there are four major category of mechanisms \cite{Zhou2012}. The first is \textit{data sample manipulation}, which works by generating different data samples to train individual learners. For example, bootstrap sampling \cite{Efron:Tibshirani1993} is exploited by Bagging \cite{Breiman1996}, whereas sequential importance sampling is adopted by AdaBoost \cite{Freund:Schapire1997}.
The second is \textit{input feature manipulation}, which works by generating different feature subspaces to train individual learners. For example, the Random Subspace approach \cite{Ho1998} randomly picks a subset of features for each individual learner. The third is \textit{learning parameter manipulation}, which works by using different parameter settings of the base learning algorithm to generate diverse individual learners. For example, different initial weights can be used for individual neural networks \cite{Kolen-NIPS-91}, whereas different split selections can be applied to individual decision trees \cite{Liu:Ting2008}. The fourth is \textit{output representation manipulation}, which works by using different output representations to generate diverse individual learners. For example, the ECOC approach \cite{Dietterich:Bakiri1995} employs error-correcting output codes, whereas the Flipping Output method \cite{Breiman2000} randomly changes the labels of some training instances. Different mechanisms can be used together, e.g., in \cite{Breiman2001,Zhou:Yu2005}. Note that, however, these mechanisms are not always effective. For example, data sample manipulation does not work well with \textit{stable learners} whose performance does not significantly change according to slight modification of training data. More information about ensemble learning can be found in \cite{Zhou2012}.

Next section will introduce the gcForest, which can be viewed as a decision tree ensemble approach that utilizes almost all categories of mechanisms for diversity enhancement.

\section{The gcForest Approach}\label{sec:approach}\vspace{-4mm}

In this section we will first introduce the cascade forest structure, and then the multi-grained scanning, followed by the overall architecture and remarks on hyper-parameters.

\subsection{Cascade Forest Structure}\label{sec:cascade}

\begin{figure}[!ht]
\centering
\includegraphics[width=.85\textwidth]{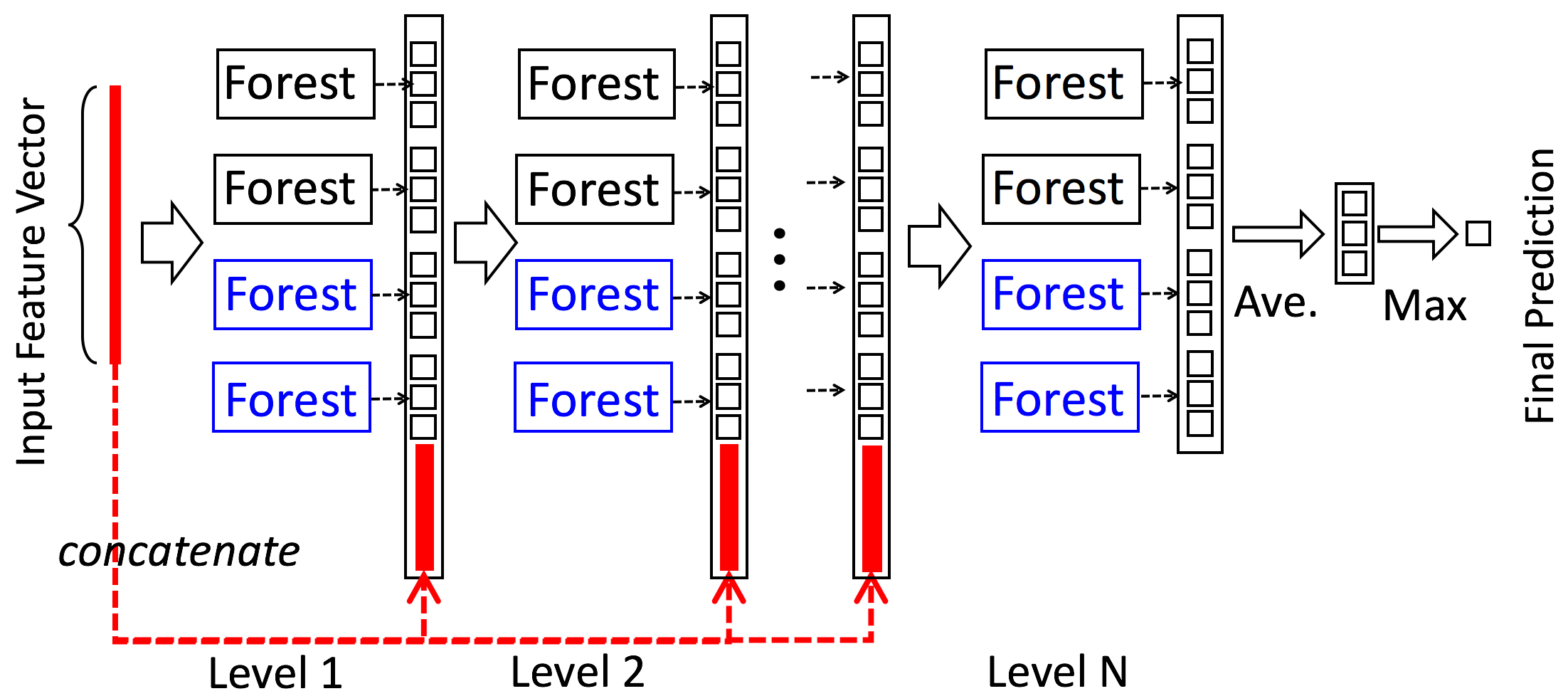}
\caption{Illustration of the cascade forest structure. Suppose each level of the cascade consists of two random forests (black) and two completely-random tree forests (blue). Suppose there are three classes to predict; thus, each forest will output a three-dimensional class vector, which is then concatenated for re-representation of the original input.}\label{fig:cascade}\bigskip
\end{figure}

Representation learning in deep neural networks mostly relies on the layer-by-layer processing of raw features. Inspired by this recognition, gcForest employs a cascade structure, as illustrated in Figure~\ref{fig:cascade}, where each level of cascade receives feature information processed by its preceding level, and outputs its processing result to the next level.

Each level is an ensemble of decision tree forests, i.e., an \textit{ensemble of ensembles}. Here, we include different types of forests to encourage the \textit{diversity}, because diversity is crucial for ensemble construction \cite{Zhou2012}. For simplicity, suppose that we use two completely-random tree forests and two random forests \cite{Breiman2001}. Each completely-random tree forest contains 500 completely-random trees \cite{Liu:Ting2008}, generated by randomly selecting a feature for split at each node of the tree, and growing tree until pure leaf, i.e., each leaf node contains only the same class of instances. Similarly, each random forest contains 500 trees, by randomly selecting $\sqrt{d}$ number of features as candidate ($d$ is the number of input features) and choosing the one with the best \textit{gini} value for split. The number of trees in each forest is a hyper-parameter, which will be discussed in Section~\ref{sec:overall}.

\begin{figure}[!b]
\centering
\includegraphics[width=0.75\textwidth]{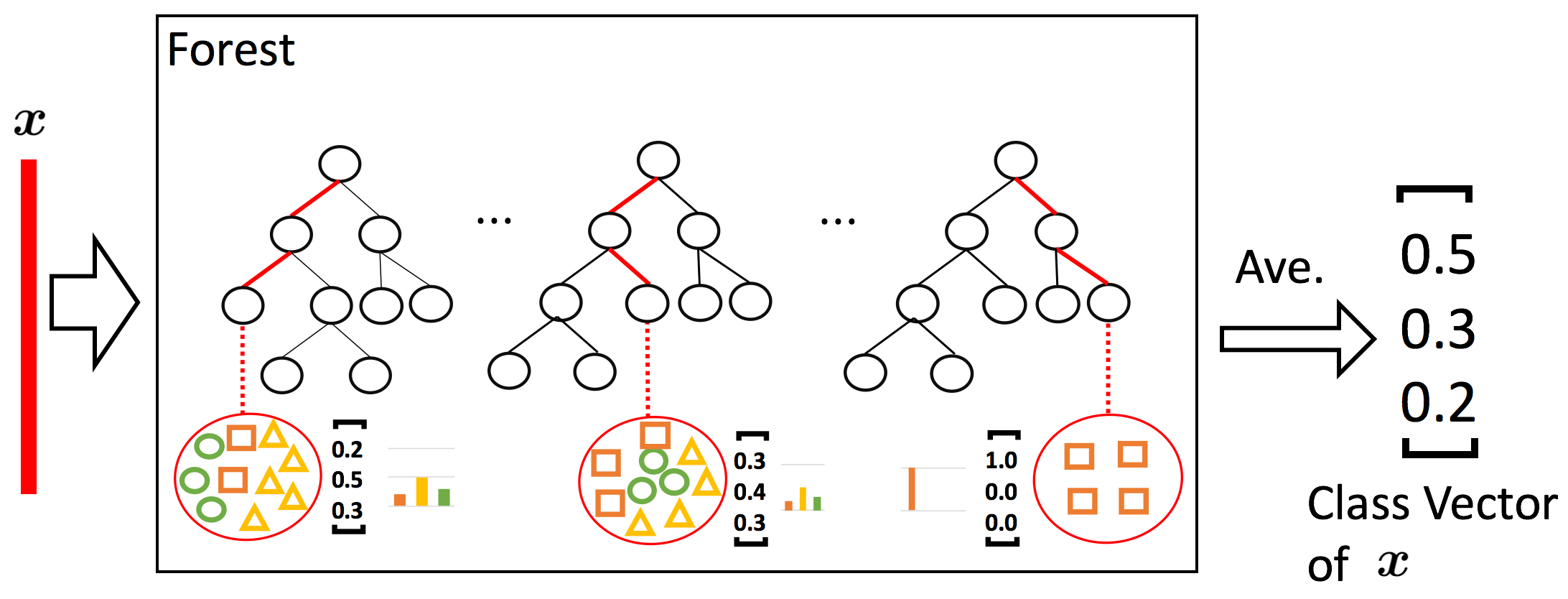}
\caption{Illustration of class vector generation. Different marks in leaf nodes imply different classes.}\label{fig:class}\bigskip
\end{figure}

Given an instance, each forest will produce an estimate of class distribution, by counting the percentage of different classes of training examples at the leaf node where the concerned instance falls, and then averaging across all trees in the same forest, as illustrated in Figure~\ref{fig:class}, where red color highlights paths along which the instance traverses to leaf nodes.

The estimated class distribution forms a class vector, which is then concatenated with the original feature vector to be input to the next level of cascade. For example, suppose there are three classes, then each of the four forests will produce a three-dimensional
class vector; thus, the next level of cascade will receive $12~(=3\times4)$ augmented features.

Note that here we take the simplest form of class vectors, i.e., the class distribution at the leaf nodes into which the concerned instance falls. It is evident that such a small number of augmented features may deliver very limited augmented information, and it is very likely to be drown out when the original feature vectors are high-dimensional. We will show in experiments that such a simple feature augmentation has already been beneficial. It is expectable that more profit can be obtained if more augmented features are involved. Actually, it is apparent that more features may be incorporated, such as class distribution of the parent nodes which express prior distribution, the sibling nodes which express complementary distribution, etc. We leave these possibilities for future exploration.

To reduce the risk of overfitting, class vector produced by each forest is generated by $k$-fold cross validation. In detail, each instance will be used as training data for $k-1$ times, resulting in $k-1$ class vectors, which are then averaged to produce the final class vector as augmented features for the next level of cascade. After expanding a new level, the performance of the whole cascade can be estimated on validation set, and the training procedure will terminate if there is no significant performance gain; thus, the number of cascade levels is automatically determined. Note that the training error rather than cross validation error can also be used to control the cascade growth when the training cost is concerned or limited computation resource available. In contrast to most deep neural networks whose model complexity is fixed, gcForest adaptively decides its model complexity by terminating training when adequate. This enables it to be applicable to different scales of training data, not limited to large-scale ones.

\subsection{Multi-Grained Scanning}\vspace{-5mm}

Deep neural networks are powerful in handling feature relationships, e.g., convolutional neural networks are effective on image data where spatial relationships among the raw pixels are critical \cite{LeCun:Bottou1998,AlexNet2012}; recurrent neural networks are effective on sequence data where sequential relationships are critical \cite{Graves:Mohamed2013,Cho:vanMerienboer2014}. Inspired by this recognition, we enhance cascade forest with a procedure of multi-grained scanning.

\begin{figure}[!t]
\centering
\includegraphics[width=.85\textwidth ]{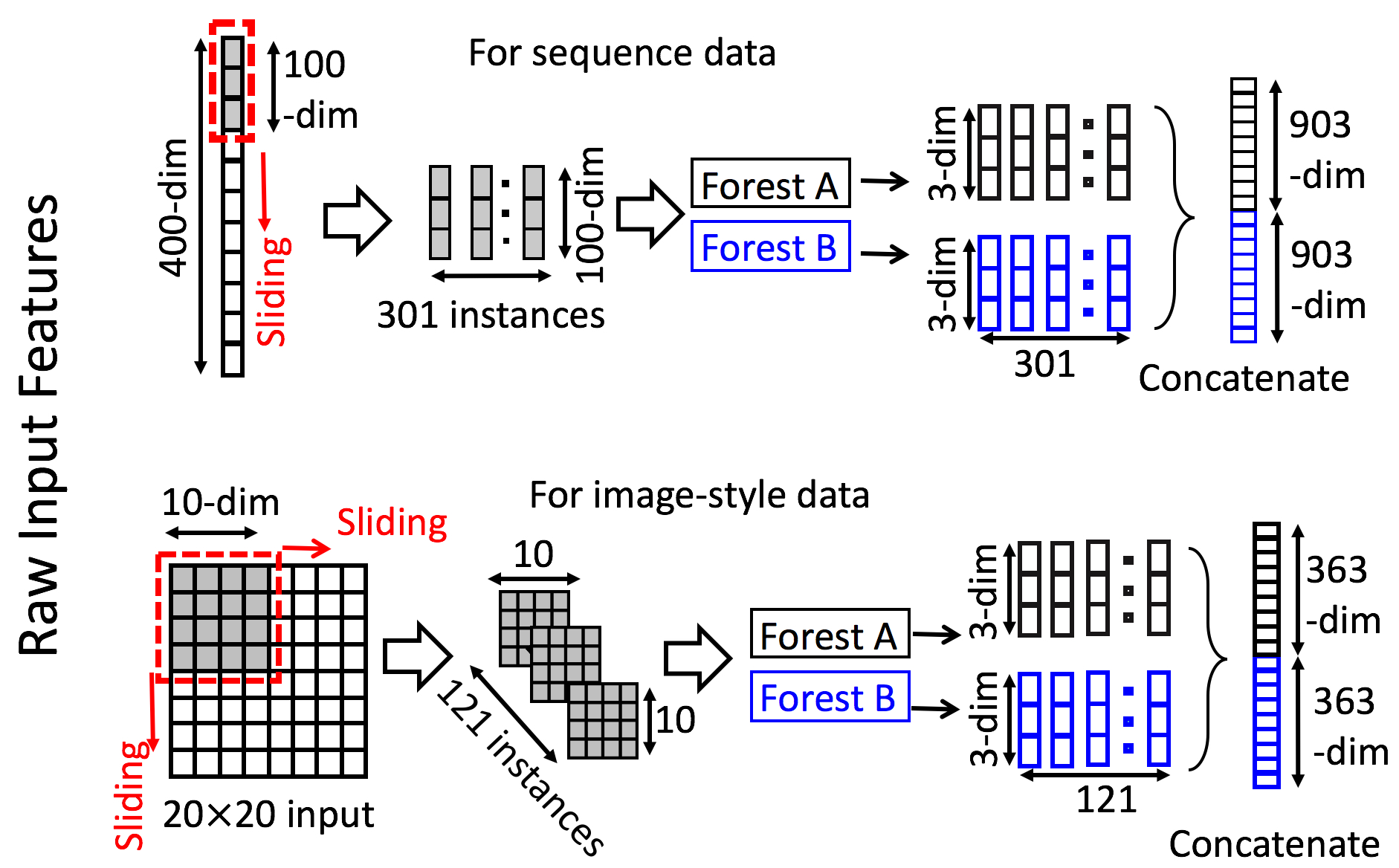}
\caption{Illustration of feature re-representation using sliding window scanning. Suppose there are three classes, raw features are 400-dim, and sliding window is 100-dim.
}\label{fig:grained}\bigskip
\end{figure}

As Figure~\ref{fig:grained} illustrates, sliding windows are used to scan the raw features. Suppose there are 400 raw features and a window size of 100 features is used. For sequence data, a 100-dimensional feature vector will be generated by sliding the window for one feature; in total 301 feature vectors are produced. If the raw features are with spacial relationships, such as a $20 \times 20$ panel of 400 image pixels, then a $10 \times 10$ window will produce 121 feature vectors (i.e., 121 $10 \times 10$ panels). All feature vectors extracted from positive/negative training examples are regarded as positive/negative instances, which will then be used to generate class vectors like in Section~\ref{sec:cascade}: the instances extracted from the same size of windows will be used to train a completely-random tree forest and a random forest, and then the class vectors are generated and concatenated as transformed features. As Figure~\ref{fig:grained} illustrates, suppose that there are 3 classes and a 100-dimensional window is used; then, 301 three-dimensional class vectors are produced by each forest, leading to a 1,806-dimensional transformed feature vector corresponding to the original 400-dimensional raw feature vector.

For the instances extracted from the windows, we simply assign them with the label of the original training example. Here, some label assignments are inherently incorrect. For example, suppose the original training example is a positive image about ``car''; it is clearly that many extracted instances do not contain a car, and therefore, they are incorrectly labeled as positive. This is actually related to the Flipping Output method \cite{Breiman2000}, a representative of output representation manipulation for ensemble diversity enhancement.

Note that when transformed feature vectors are too long to be accommodated, feature sampling can be performed, e.g., by subsampling the instances generated by sliding window scanning, since completely-random trees do not rely on feature split selection whereas random forests are quite insensitive to inaccurate feature split selection. Such a feature sampling process is also related to the Random Subspace method \cite{Ho1998}, a representative of input feature manipulation for ensemble diversity enhancement.

\begin{landscape}
\begin{figure}
\centering
\includegraphics[width=1.65\textwidth]{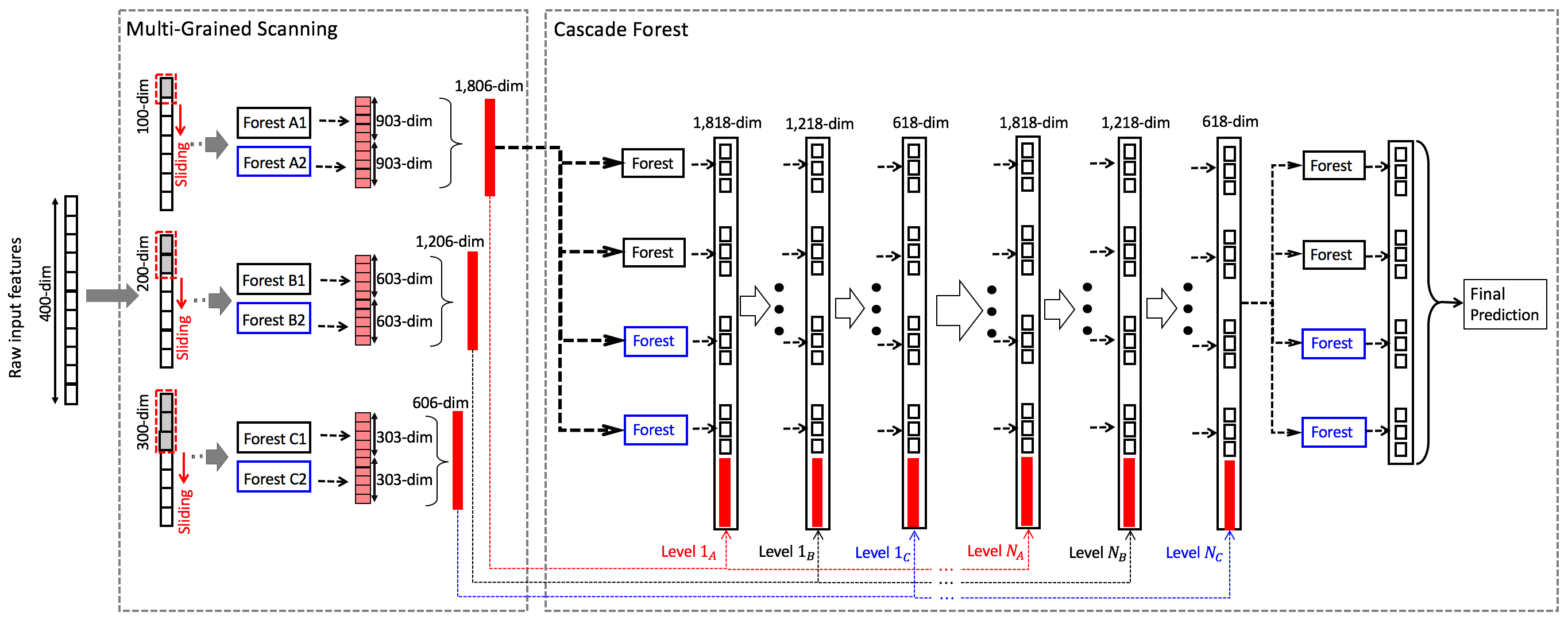}
\caption{The overall procedure of gcForest. Suppose there are three classes to predict, raw features are 400-dim, and three sizes of sliding windows are used.}\label{fig:overall}
\end{figure}
\end{landscape}

Figure~\ref{fig:grained} shows only one size of sliding window. By using multiple sizes of sliding windows, differently grained feature vectors will be generated, as shown in Figure~\ref{fig:overall}.

\subsection{Overall Procedure and Hyper-Parameters}\label{sec:overall}\vspace{-5mm}

Figure~\ref{fig:overall} summarizes the overall procedure of gcForest. Suppose that the original input is of 400 raw features, and three window sizes are used for multi-grained scanning. For $m$ training examples, a window with size of 100 features will generate a data set of $301 \times m$ 100-dimensional training examples. These data will be used to train a completely-random tree forest and a random forest, each containing 500 trees. If there are three classes to be predicted, a 1,806-dimensional feature vector will be obtained as described in Section~\ref{sec:cascade}. The transformed training set will then be used to train the 1st-grade of cascade forest.

Similarly, sliding windows with sizes of 200 and 300 features will generate 1,206-dimensional and 606-dimensional feature vector, respectively, for each original training example. The transformed feature vectors, augmented with the class vector generated by the previous grade, will then be used to train the 2nd-grade and 3rd-grade of cascade forests, respectively. This procedure will be repeated till convergence of validation performance. In other words, the final model is actually a \textit{cascade of cascades}, where  each cascade consists of multiple levels each corresponding to a grain of scanning, e.g., the 1st cascade consists of Level $1_{A}$ to Level $1_{C}$, as shown in Figure~\ref{fig:overall}. Note that for difficult tasks, users can try more grains if computational resource allows.

Given a test instance, it will go through the multi-grained scanning procedure to get its corresponding transformed feature representation, and then go through the cascade till the last level. The final prediction will be obtained by aggregating the four 3-dimensional class vectors at the last level, and taking the class with the maximum aggregated value.

Table~\ref{tab:parameter} summarizes the hyper-parameters of deep neural networks and gcForest, where the default values used in our experiments are given.

\renewcommand{\baselinestretch}{0.95}
\small\normalsize

\begin{landscape}
\begin{table}[!ht]
\centering
\caption{Summary of hyper-parameters and default settings. Boldfont highlights hyper-parameters with relatively larger influence; ``?'' indicates default value unknown, or generally requiring different settings for different tasks. }\smallskip\smallskip\label{tab:parameter}
\begin{tabular}{l|l|}
\hline
\multicolumn{1}{|l|}{Deep neural networks (e.g., convolutional neural networks)}   & gcForest                                           \\ \hline
\multicolumn{1}{|l|}{Type of activation functions:}                & Type of forests:   \\
\multicolumn{1}{|l|}{~~~~Sigmoid, ReLU, tanh, linear, etc.}        & ~~~~Completely-random tree forest, random forest, etc. \\
\multicolumn{1}{|l|}{Architecture configurations:}                 & Forest in multi-grained scanning:    \\
\multicolumn{1}{|l|}{~~~~ \textbf{No. Hidden layers}: ?}           & ~~~~\textbf{No. Forests}: \{2\}               \\
\multicolumn{1}{|l|}{~~~~ \textbf{No. Nodes in hidden layer}: ?}   & ~~~~\textbf{No. Trees in each forest}: \{500\}  \\
\multicolumn{1}{|l|}{~~~~ \textbf{No. Feature maps}: ?}            & ~~~~Tree growth: till pure leaf, or reach depth 100 \\
\multicolumn{1}{|l|}{~~~~~\textbf{Kernel size}: ?}                 & ~~~~\textbf{Sliding window size}: \{$\lfloor d/16 \rfloor$, $\lfloor d/8 \rfloor$, $\lfloor d/4 \rfloor$\}  \\

\multicolumn{1}{|l|}{Optimization configurations:}                 & Forest in cascade:  \\
\multicolumn{1}{|l|}{~~~~ \textbf{Learning rate}: ?}               & ~~~~\textbf{No. Forests}: \{8\}           \\
\multicolumn{1}{|l|}{~~~~ Dropout: \{0.25/0.50\}}                      & ~~~~\textbf{No. Trees in each forest}: \{500\} \\
\multicolumn{1}{|l|}{~~~~ \textbf{Momentum}: ?}                    & ~~~~Tree growth: till pure leaf \\
\multicolumn{1}{|l|}{~~~~ \textbf{L1/L2 weight regularization penalty}: ?}     &      \\
\multicolumn{1}{|l|}{~~~~~Weight initialization: Uniform, glorot\_normal, glorot\_uni, etc. }        &      \\
\multicolumn{1}{|l|}{~~~~~Batch size: \{32/64/128\}}                     &         \\\hline
\end{tabular}
\end{table}
\end{landscape}

\renewcommand{\baselinestretch}{1.4}
\small\normalsize

\section{Experiments}\label{sec:experiments}\vspace{-4mm}

\subsection{Configuration}\vspace{-5mm}

In this section we compare gcForest with deep neural networks and several other popular learning algorithms. The goal is to validate that gcForest can achieve performance highly competitive to deep neural networks, with easier parameter tuning even across a variety of tasks.

In all experiments gcForest is using the \textit{same} cascade structure: Each level consists of 4 completely-random tree forests and 4 random forests, each containing 500 trees, as described in Section~\ref{sec:cascade}. Three-fold cross validation is used for class vector generation. The number of cascade levels is automatically determined. In detail, we split the training set into two parts, i.e., growing set and estimating set\footnote{Some experimental datasets are given with training/validation sets. To avoid confusion, here we call the subsets generated from training set as growing/estimating sets.}; then we use the growing set to grow the cascade, and the estimating set to estimate the performance. If growing a new level does not improve the performance, the growth of the cascade terminates and the estimated number of levels is obtained. Then, the cascade is retrained based on merging the growing and estimating sets. For all experiments we take 80\% of the training data for growing set and 20\% for estimating set. For multi-grained scanning, three window sizes are used. For $d$ raw features, we use feature windows with sizes of $\lfloor d/16 \rfloor$, $\lfloor d/8 \rfloor$, $\lfloor d/4 \rfloor$; if the raw features are with panel structure (such as images), the feature windows are also with panel structure as shown in Figure~\ref{fig:grained}. Note that a careful task-specific tuning may bring better performance; here, to highlight that the hyper-parameter setting of gcForest is much easier than deep neural networks, we simply use the same setting for all tasks, whereas task-specific tunings are performed for DNNs.

For deep neural network configurations, we use ReLU for activation function, cross-entropy for loss function, adadelta for optimization, dropout rate 0.25 or 0.5 for hidden layers according to the scale of training data. The network structure hyper-parameters, however, cannot be fixed across tasks, otherwise the performance will be embarrassingly unsatisfactory. For example, a network attained 80\% accuracy on ADULT dataset achieved only 30\% accuracy on YEAST with the same architecture (only the number of input/output nodes changed to suit the data). Therefore, for deep neural networks, we examine a variety of architectures on validation set, and pick the one with the best performance, then re-train the whole network on training set and report the test accuracy.

\subsection{Results}\vspace{-5mm}

We run experiments on a broad range of tasks.

\subsubsection*{Image Categorization}

The MNIST dataset \cite{LeCun:Bottou1998} contains 60,000 images of size $28$ by $28$ for training (and validating), and 10,000 images for testing. We compare it with a re-implementation of LeNet-5 (a modern version of LeNet with dropout and ReLUs), SVM with rbf kernel, and a standard Random Forest with 2,000 trees. We also include the result of the Deep Belief Nets reported in \cite{hinton2006fast}. The test results are summarized in Table~\ref{tab:MNIST}, showing that gcForest, although simply using default settings in Table~\ref{tab:parameter}, achieves highly competitive performance.\smallskip\smallskip

\renewcommand{\baselinestretch}{0.95}
\small\normalsize

\begin{table}[!ht]
\centering
\caption{Comparison of test accuracy on MNIST}\label{tab:MNIST}\smallskip\smallskip
\begin{tabular}{|l|l|}
\hline
\textbf{gcForest}      & \textbf{99.26\%}  \\ \hline
LeNet-5                       & 99.05\%  \\ \hline
Deep Belief Net               & 98.75\% {\small \cite{hinton2006fast}}  \\ \hline
SVM {\small (rbf kernel)}     & 98.60\%   \\ \hline
Random Forest                 & 96.80\%   \\ \hline
\end{tabular}
\end{table}

\renewcommand{\baselinestretch}{1.4}
\small\normalsize


\subsubsection*{Face Recognition}

The ORL dataset \cite{Samaria:Harter1994} contains 400 gray-scale facial images taken from 40 persons. We compare it with a CNN consisting of 2 conv-layers with 32 feature maps of $3\times3$ kernel, and each conv-layer has a $2\times2$ max-pooling layer followed. A dense layer of 128 hidden units is fully connected with the convolutional layers and finally a fully connected soft-max layer with 40 hidden units is appended at the end. ReLU, cross-entropy loss, dropout rate of 0.25 and adadelta are used for training. The batch size is set to 10, and 50 epochs are used. We have also tried other configurations of CNN, whereas this one gives the best performance. We randomly choose 5/7/9 images per person for training, and report the test performance on the remaining images. Note that a random guess will achieve 2.5\% accuracy, since there are 40 possible outcomes. The $k$NN method here uses $k=3$ for all cases. The test results are summarized in Table~\ref{tab:ORL}.
\footnote{There are studies where CNNs perform more excellently for face recognition, by using huge amount of face images to train the model. Here, we simply use the training data.}
The gcForest runs well across all three cases even by using the same configurations as described in Table~\ref{tab:parameter}.\smallskip\smallskip

\renewcommand{\baselinestretch}{0.95}
\small\normalsize

\begin{table}[!ht]
\centering
\caption{Comparison of test accuracy on ORL}\label{tab:ORL}\smallskip\smallskip
\begin{tabular}{|l|c|c|c|}
\hline
                            & 5 image  & 7 images & 9 images \\ \hline
\textbf{gcForest}  & \textbf{91.00\%}  & \textbf{96.67\% }  & \textbf{97.50\%}   \\ \hline
Random Forest               & 91.00\%            & 93.33\%            & 95.00\%  \\ \hline
CNN                         & 86.50\%            & 91.67\%            & 95.00\%  \\ \hline
SVM {\small(rbf kernel)}    & 80.50\%            & 82.50\%            & 85.00\%  \\ \hline
$k$NN                       & 76.00\%            & 83.33\%            & 92.50\%   \\ \hline
\end{tabular}
\end{table}

\renewcommand{\baselinestretch}{1.4}
\small\normalsize

\subsubsection*{Music Classification}

The GTZAN dataset \cite{Tzanetakis:Cook2002} contains 10 genres of music clips, each represented by 100 tracks of 30 seconds long. We split the dataset into 700 clips for training and 300 clips for testing. In addition, we use MFCC feature to represent each 30 seconds music clip, which transforms the original sound wave into a $1,280\times13$ feature matrix.
Each frame is atomic according to its own nature; thus, CNN uses a $13\times8$ kernel with 32 feature maps as the conv-layer, each followed by a pooling layer. Two fully connected layers with 1,024 and 512 units, respectively, are appended, and finally a soft-max layer is added in the last. We also compare it with an MLP having two hidden layers, with 1,024 and 512 units, respectively. Both networks use ReLU as activation function and categorical cross-entropy as the loss function. For Random Forest, Logistic Regression and SVM, each input is concatenated into an $1,280\times13$ feature vector. The test results are summarized in Table~\ref{tab:GTZAN}.\smallskip\smallskip

\renewcommand{\baselinestretch}{0.95}
\small\normalsize

\begin{table}[!ht]
\centering
\caption{Comparison of test accuracy on GTZAN}\label{tab:GTZAN}\smallskip\smallskip
\begin{tabular}{|l|l|}
\hline
\textbf{gcForest}           & \textbf{65.67}\%       \\ \hline
CNN                         & 59.20\%   \\ \hline
MLP                         & 58.00\%     \\ \hline
Random Forest               & 50.33\%  \\ \hline
Logistic Regression         & 50.00\%   \\ \hline
SVM (rbf kernel)            & 18.33\%    \\ \hline
\end{tabular}\bigskip
\end{table}

\renewcommand{\baselinestretch}{1.4}
\small\normalsize


\subsubsection*{Hand Movement Recognition}

The sEMG dataset \cite{Sapsanis:Georgoulas2013} consists of 1,800 records each belonging to one of six hand movements, i.e.,
spherical, tip, palmar, lateral, cylindrical and hook.
This is a time-series dataset, where EMG sensors capture 500 features per second and each record associated with 3,000 features. In addition to an MLP with \textit{input}-1,024-512-\textit{output} structure, we also evaluate a recurrent neural network, LSTM \cite{Gers:Eck:Schmidhuber2001} with 128 hidden units and sequence length of 6 (500-dim input vector per second).
The test results are summarized in Table~\ref{tab:sEMG}. \smallskip\smallskip

\renewcommand{\baselinestretch}{0.95}
\small\normalsize

\begin{table}[!ht]
\centering
\caption{Comparison of test accuracy on sEMG data}\label{tab:sEMG}\smallskip\smallskip
\begin{tabular}{|l|l|}
\hline
\textbf{gcForest}           & \textbf{71.30}\%       \\ \hline
LSTM                         & 45.37\%     \\ \hline
MLP                         & 38.52\%     \\ \hline
Random Forest               & 29.62\%  \\ \hline
SVM (rbf kernel)            & 29.62\%    \\ \hline
Logistic Regression         & 23.33\%   \\ \hline
\end{tabular}
\end{table}

\renewcommand{\baselinestretch}{1.4}
\small\normalsize

\subsubsection*{Sentiment Classification}

The IMDB dataset \cite{Maas:Daly:Pham2011} contains 25,000 movie reviews for training and 25,000 for testing. The reviews are represented by \textit{tf-idf} features. This is not image data, and thus CNNs are not directly applicable. So, we compare it with an MLP with structure \textit{input}-1,024-1,024-512-256-\textit{output}. We also include the result reported in \cite{kim2014convolutional}, which uses CNNs facilitated with word embeding. Considering that \textit{tf-idf} features do not convey spacial or sequential relationships, we skip multi-grained scanning for gcForest. The test accuracy is summarized in Table~\ref{tab:IMDB}.\smallskip\smallskip

\renewcommand{\baselinestretch}{0.95}
\small\normalsize

\begin{table}[!ht]
\centering
\caption{Comparison of test accuracy on IMDB}\label{tab:IMDB}\smallskip\smallskip
\begin{tabular}{|l|l|}
\hline
\textbf{gcForest}          & \textbf{89.16\%}  \\ \hline
CNN                        & 89.02\% {\small \cite{kim2014convolutional}} \\ \hline
MLP                        & 88.04\%  \\ \hline
Logistic Regression        & 88.62\%  \\ \hline
SVM {\small (linear kernel)}  & 87.56\%   \\ \hline
Random Forest              & 85.32\%  \\ \hline
\end{tabular}
\end{table}

\renewcommand{\baselinestretch}{1.4}
\small\normalsize

\subsection{Low-Dimensional Data}\vspace{-5mm}

We also evaluate gcForest on UCI-datasets \cite{Lichman2013} with relatively small number of features: LETTER with 16 features and 16,000/4,000 training/test examples, ADULT with 14 features and 32,561/16,281 training/test examples, and YEAST with only 8 features and 1,038/446 training/test examples. Fancy architectures like CNNs could not work on such data as there are too few features without spatial relationship. So, we compare it with MLPs. Unfortunately, although MLPs have less configuration options than CNNs, they are still very tricky to set up. For example, MLP with \textit{input}-16-8-8-\textit{output} structure and ReLU activation achieve 76.37\% accuracy on ADULT but just 33\% on LETTER. We conclude that there is no way to pick one MLP structure which gives decent performance across all datasets. Therefore, we report different MLP structures with the best performance: for LETTER the structure is \textit{input}-70-50-\textit{output}, for ADULT is \textit{input}-30-20-\textit{output}, and for YEAST is \textit{input}-50-30-\textit{output}. In contrast, gcForest uses the same configuration as shown in Table~\ref{tab:parameter}, except that the multi-grained scanning is abandoned considering that the features of these small-scale data do not hold spacial or sequential relationships.
The test results are summarized in Table~\ref{tab:uci_result}.\smallskip\smallskip

\renewcommand{\baselinestretch}{0.95}
\small\normalsize

\begin{table}[!ht]
\centering
\caption{Comparison of test accuracy on low-dim data}\label{tab:uci_result}\smallskip\smallskip
\begin{tabular}{|l|c|c|c|}
\hline
     & LETTER & ADULT & YEAST  \\ \hline
\textbf{gcForest}   & \textbf{97.40\%}      &   \textbf{86.40\%}    & \textbf{63.45\%}   \\ \hline
Random Forest       & 96.50\%       &  85.49\%      & 61.66\%         \\ \hline
MLP                 & 95.70\%       &  85.25\%      & 55.60\%      \\ \hline
\end{tabular}
\end{table}

\renewcommand{\baselinestretch}{1.4}
\small\normalsize

\subsection{High-Dimensional Data}\vspace{-5mm}

The CIFAR-10 dataset \cite{krizhevsky2009learning} contains 50,000 images of 10 classes for training and 10,000 images for testing. Here, each image is a 32 by 32 colored image with 8 gray-levels; thus, each instance is of 8192-dim. The test results are shown in Table~\ref{tab:CIFAR10}, which also includes results of several deep neural networks reported in literature.

\renewcommand{\baselinestretch}{0.95}
\small\normalsize

\begin{table}[!ht]
\centering
\caption{Comparison of test accuracy on CIFAR-10.}\label{tab:CIFAR10}\smallskip\smallskip
\begin{tabular}{|l|l|}
\hline
ResNet                           & 93.57\% {\small \cite{he2016deep}} \\ \hline
AlexNet                          & 83.00\% {\small \cite{AlexNet2012}} \\ \hline
\textbf{gcForest(gbdt)}          & \textbf{69.00\%}  \\ \hline
\textbf{gcForest(5grains)}       & \textbf{63.37\%}  \\ \hline
Deep Belief Net                  & 62.20\%  {\small \cite{krizhevsky2009learning} } \\ \hline
\textbf{gcForest(default)}       & \textbf{61.78\%}  \\ \hline
Random Forest                    & 50.17\%  \\ \hline
MLP                              & 42.20\% {\small \cite{ba2014deep}} \\ \hline
Logistic Regression              & 37.32\% \\ \hline
SVM {\small (linear kernel)}     & 16.32\% \\ \hline
\end{tabular}\bigskip
\end{table}

\renewcommand{\baselinestretch}{1.4}
\small\normalsize

As we discussed in Section~\ref{sec:approach}, currently we only use 10-dim augmented feature vector from each forest, and such a small number of augmented features will be easily drown out in the original long feature vector. Nevertheless, although the gcForest with default setting, i.e., gcForest(default), is inferior to state-of-the-art DNNs, it is already the best among non-DNN approaches. Moreover, the performance of gcForest can be further improved via task-specific tuning, e.g., by including more grains (i.e., using more sliding window sizes in multi-grained scanning) like gcForest(5grains) which uses five grains. It is also interesting to see that the performance gets significant improvement with gcForest(gbdt), which simply replaces the final level with GBDT \cite{Chen2016xgboost}. Section~\ref{sec:larger} will show that better performance can be obtained if larger models of gcForest can be trained.

\subsection{Running time}\vspace{-5mm}

Our experiments use a PC with 2 Intel E5 2695 v4 CPUs (18 cores), and the running efficiency of gcForest is good. For example, for IMDB dataset (25,000 examples with 5,000 features), it takes 267.1 seconds per cascade level, and automatically terminates with 9 cascade levels, amounting to 2,404 seconds or 40 minutes. In contrast, MLP compared on the same dataset requires 50 epochs for convergence and 93 seconds per epoch, amounting to 4,650 seconds or 77.5 minutes for training; 14 seconds per epoch (with batch size of 32) if using GPU (Nvidia Titan X pascal), amounting to 700 seconds or 11.6 minutes. Multi-grained scanning will increase the cost of gcForest; however, the different grains of scanning are inherently parallel. Also, both completely-random tree forests and random forests are parallel ensemble methods \cite{Zhou2012}. Thus, the efficiency of gcForest can be improved further with optimized parallel implementation. Note that the training cost is controllable because users can set the number of grains, forests, trees by considering computational cost available. It is also noteworthy that the above comparison is somewhat unfair to gcForest, because many different architectures have been tried for neural networks to achieve the reported performance but these time costs are not included.

\subsection{Influence of Multi-Grained Scanning}\vspace{-5mm}

To study the separate contribution of the cascade forest structure and multi-grained scanning, Table~\ref{tab:ingredient} compares gcForest with cascade forest on MNIST,  GTZAN and sEMG datasets. It is evident that when there are spacial or sequential feature relationships, the multi-grained scanning process helps improve performance apparently.\smallskip\smallskip

\renewcommand{\baselinestretch}{0.95}
\small\normalsize

\begin{table}[!ht]
\centering
\caption{Results of gcForest w/wo multi-grained scanning}\label{tab:ingredient}\smallskip\smallskip
\begin{tabular}{|l|c|c|c|}
\hline
                &   MNIST   & GTZAN     & sEMG   \\ \hline
gcForest        &  \textbf{99.26\%}  & \textbf{65.67\%}   &  \textbf{71.30\%}     \\ \hline
CascadeForest   &  98.02\%  & 52.33\%   &  48.15\%     \\ \hline
\end{tabular}
\end{table}

\renewcommand{\baselinestretch}{1.4}
\small\normalsize

\subsection{Influence of Cascade Structure}\vspace{-5mm}

\begin{landscape}
\begin{figure}
\centering
\includegraphics[width=1.65\textwidth]{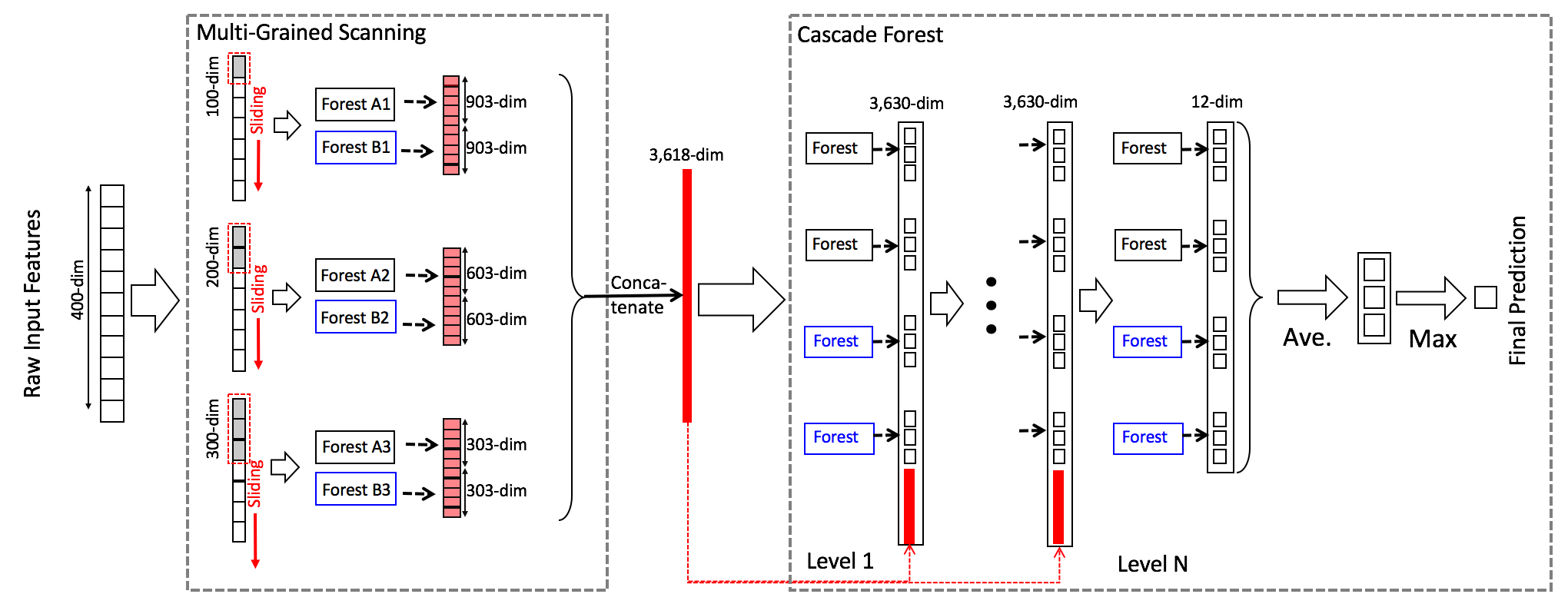}
\caption{The variant gcForest$_{conc}$ which concatenates all features from multiple grains. Suppose there are three classes to predict, raw features are 400-dim, and three sizes of sliding windows are used.}\label{fig:overallA}
\end{figure}
\end{landscape}

The final model structure of gcForest is a \textit{cascade of cascades}, where each cascade consists of multiple levels each corresponding to a grain of scanning, as shown in Figure~\ref{fig:overall}. There are other possible ways to exploit the features from multiple grains, e.g., by concatenating all the features together, as shown in Figure~\ref{fig:overallA}.

Table~\ref{tab:concat}\footnote{Here, ORL is with 9 training images per person.} compares gcForest with the gcForest$_{conc}$, which shows that concatenating the features from multiple grains is not as good as the current design in gcForest. Nevertheless, there might be other ways leading to better results; we leave it for future exploration.\smallskip\smallskip

\renewcommand{\baselinestretch}{0.95}
\small\normalsize

\begin{table}[!ht]
\centering
\caption{Results of gcForest with the variant of concatenating features from multiple grains.}\label{tab:concat}\smallskip\smallskip
\scriptsize
\begin{tabular}{|l|c|c|c|c|c|c|c|c|}
\hline
                     &   MNIST   & ORL & GTZAN     & sEMG     & IMDB     & LETTER & ADULT & YEAST \\ \hline
gcForest            &  \textbf{99.26\%}  & 97.50\% & 65.67\%   &  \textbf{71.30\%}  & 89.16\%  & \textbf{97.40\%} & \textbf{86.40\%} & \textbf{63.45\%}\\ \hline
gcForest$_{conc}$   &  98.96\%  & \textbf{98.30\%} & 65.67\%   &  55.93\%  & \textbf{89.32\%}  & 97.25\% & 86.17\% & 63.23\%\\ \hline
\end{tabular}
\end{table}

\renewcommand{\baselinestretch}{1.4}
\small\normalsize

\subsection{Influence of Larger Models}\label{sec:larger}\vspace{-5mm}

\begin{figure}[!h]
\centering
\begin{minipage}[l]{2.5in}
\centering
\includegraphics[width=2.3in]{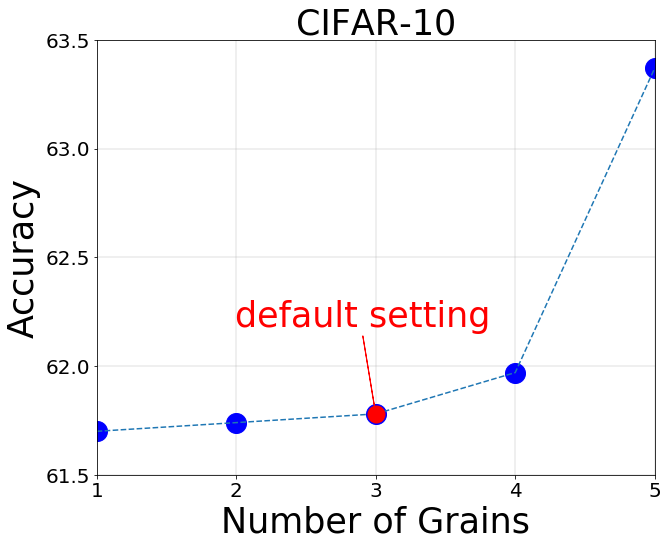}
\end{minipage}%
\begin{minipage}[l]{2.5in}
\centering
\includegraphics[width=2.3in]{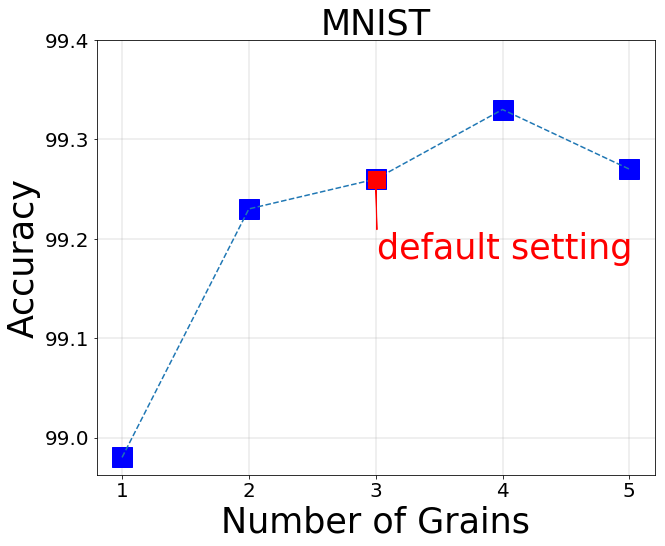}
\end{minipage}\\[+8pt]
\centering
{\small (a) With increasing number of grains.}\\[+12pt]
\centering
\begin{minipage}[l]{2.5in}
\centering
\includegraphics[width=2.3in]{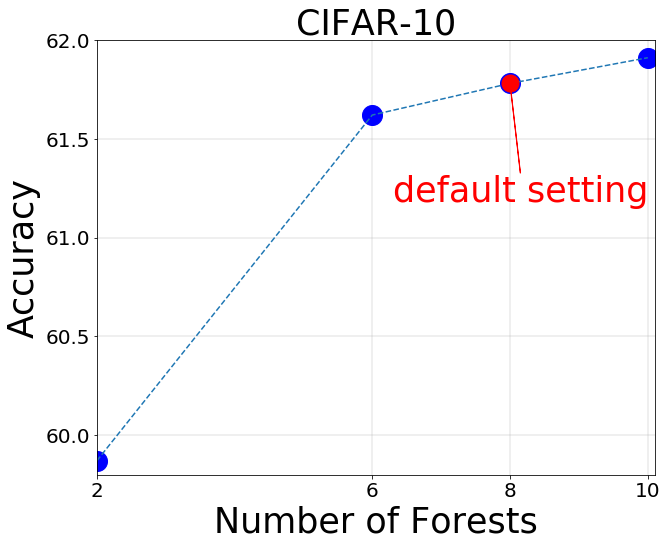}
\end{minipage}%
\begin{minipage}[l]{2.5in}
\centering
\includegraphics[width=2.32in]{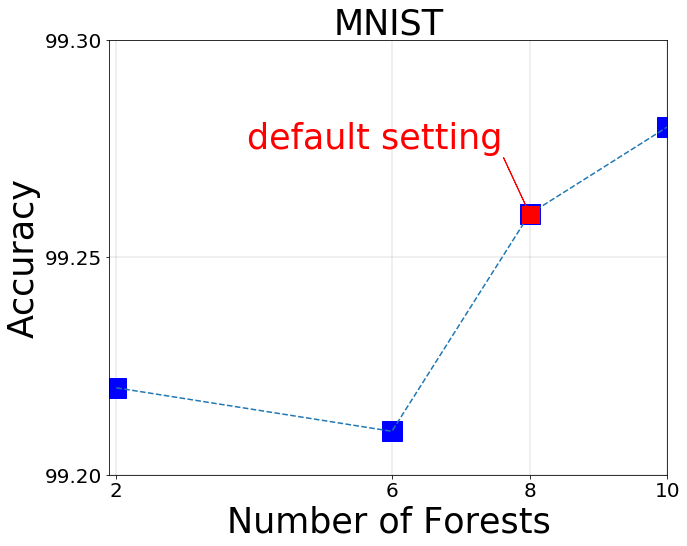}
\end{minipage}\\[+8pt]
{\small (b) With increasing number of forests per grade.}\\[+12pt]
\centering
\begin{minipage}[l]{2.5in}
\centering
\includegraphics[width=2.3in]{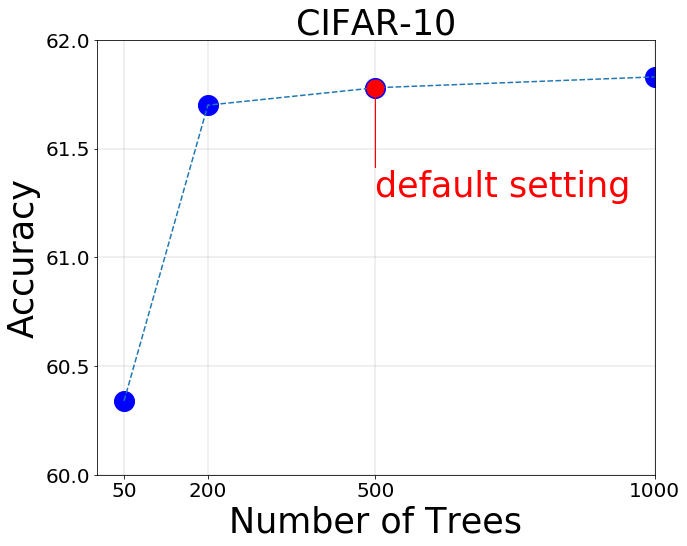}
\end{minipage}%
\begin{minipage}[l]{2.5in}
\centering
\includegraphics[width=2.3in]{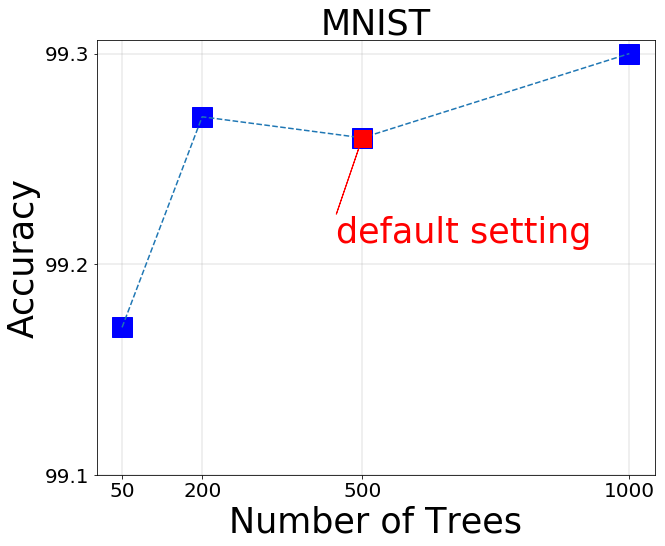}
\end{minipage}\\[+8pt]
{\small (c) With increasing number of trees per forest.}\\[+8pt]
\caption{Test accuracy of gcForest with increasing number of grains/forests/trees. Red color highlights the performance achieved by default setting.}\label{fig:apdx-fig}\bigskip
\end{figure}

Our results in Figure~\ref{fig:apdx-fig} suggest that larger models might tend to offer better performances, although we have not tried even more grains, forests and trees due to limitation of computational resource.

Note that computational facilities are crucial for enabling the training of larger models; e.g., GPUs for DNNs. On one hand, some new computational devices, such as Intel KNL of the MIC (Many Integrated Core) architecture, might offer potential acceleration for gcForest like GPUs for DNNs. On the other hand, some components of gcForest, e.g., the multi-grained scanning, may be accelerated by exploiting GPUs. Moreover, there is plenty of room for improvement with distributed computing implementations.

\section{Related Work}\label{sec:related}\vspace{-4mm}

The gcForest is a decision tree ensemble approach. Ensemble methods \cite{Zhou2012} are a kind of powerful machine learning techniques which combine multiple learners for the same task. Actually there are some studies showing that by using ensemble methods such as random forest facilitated with deep neural network features, the performance can be even better than simply using deep neural networks \cite{kontschieder2015deep}. Our purpose of using ensemble, however, is quite different. We are aiming at a non-NN style deep model rather than a combination with deep neural networks. In particular, by using the cascade forest structure, we hope to endow the model with characteristics of layer-by-layer processing, in-model feature transformation and sufficient model complexity.

Random forest \cite{Breiman2001}, which has been widely applied to various tasks, is one of the most successful ensemble methods \cite{Zhou2012}. Completely-random tree forest has been found useful during recent years, such as iForest \cite{Liu:Ting:Zhou2008} for anomaly detection, sencForest \cite{Mu:Ting:Zhou2017} for handling emerging new classes in streaming data, etc. The gcForest offers another example exhibiting the usefulness of completely-random tree forest.

Many works try to connect random forest with neural networks, such as converting cascaded random forests to convolutional neural networks \cite{Richmond:Kainmueller:Yang:Myers2015}, exploiting random forests to help initialize neural networks \cite{Welbl2014}, etc. These work are typically based on early studies connecting trees with neural networks, e.g., mapping of trees to networks \cite{Sethi1990}, tree-structured neural networks \cite{Sanger1991}, as reviewed in \cite{Zhou:Chen2002}. Their goals are totally different from ours. In particular, their final models are based on differentiable modules, whereas we are trying to develop deep models based on non-differentiable modules.

The multi-grained scanning procedure of gcForest uses different sizes of sliding windows to examine the data; this is somewhat related to wavelet and other multi-resolution examination procedures \cite{Mallat1999}. For each window size, a set of instances are generated from one training example; this is related to bag generators \cite{Wei:Zhou2016} of multi-instance learning \cite{Dietterich:Lathrop:Lozano1997}. In particular, the bottom part of Figure~\ref{fig:grained}, if applied to images, can be regarded as the \textit{SB} image bag generator \cite{Maron:Lozano1998,Wei:Zhou2016}.

The cascade procedure of gcForest is related to Boosting \cite{Freund:Schapire1997}, which is able to automatically decide the number of learners in ensemble, and particularly, a cascade boosting procedure \cite{Viola:Jones2001} has achieved great success in object detection tasks. Note that when multiple grains are used, each level in the cascade of gcForest consists of multiple grades; this is actually a cascade of cascades. Each grade can be regarded as an ensemble of ensembles. In contrast to previous studies about ensemble of ensembles, such as using Bagging as base learners for Boosting \cite{Webb2000}, gcForest uses the ensembles in the same grade together for feature re-representation.

Passing the output of one grade of learners as input to another grade of learners is related to stacking \cite{Wolpert1992,Breiman1996stack}. Based on suggestions from studies about stacking \cite{Ting:Witten1999,Zhou2012}, we use cross-validation procedure to generate inputs from one grade for the next. Note that stacking is easy to overfit with more than two grades, and cannot enable a deep model by itself.

To construct a good ensemble, it is well known that individual learners should be accurate and diverse, yet there is no well accepted formal definition of diversity \cite{Kuncheva:Whitaker2003,Zhou2012}. Thus, researchers usually try to enhance diversity heuristically, such as what we have done by using different types of forests in each grade. Actually, gcForest exploits all the four major categories of diversity enhancement mechanisms \cite{Zhou2012}.

As a tree-based approach, gcForest might be potentially easier for theoretical analysis than deep neural networks, although this is beyond the scope of this paper. Indeed, some recent theoretical studies about deep learning, e.g., \cite{Mhaskar:Liao:Poggio2017}, seem more intimate with tree-based models.

\section{Future Issues}\label{sec:future}\vspace{-4mm}

One important future issue is to enhance the feature re-representation process. The current implementation of gcForest takes the simplest form of class vectors, i.e., the class distribution at the leaf nodes into which the concerned instance falls. Such a small number of augmented features will be easily drown out when the original feature vectors are high-dimensional. It is apparent that more features may be involved, such as class distribution of the parent nodes which express prior distribution, the sibling nodes which express complementary distribution, the decision path encoding, etc. Intuitively, more features may enable the incorporation of more information, although not always necessarily helpful for generalization. Moreover, a longer class vector may enable a joint multi-grained scanning process, leading to more flexibility of re-representation. Recently we show that decision tree forests can serve as AutoEncoder \cite{Feng:Zhou2018}. On one hand, this shows that the ability of AutoEncoder is not a special property of neural networks as it had been thought before; on the other hand, it discloses that a forest can encode abundant information, and thus offers great potential to facilitate rich feature re-representation.

Another important future issue is to accelerate and reduce the memory consumption. As suggested in Section~\ref{sec:larger}, building larger deep forests may lead to better generalization performance in practice, whereas computational facilities are crucial for enabling the training of larger models. Actually, the success of DNNs owes much to the acceleration offered by GPUs, but unfortunately forest structure is not naturally suitable to GPUs. One possibility is to consider some new computational devices, such as Intel KNL of the MIC (Many Integrated Core) architecture; another is to use distributed computing implementations. Feature sampling can be executed when transformed feature vectors produced by multi-grained scanning are too long to be accommodated; this not only helps reduce storage, but also offers another channel to enhance the diversity of the ensembles. It is somewhat like combining random tree forest with random subspace \cite{Ho1998}, another powerful ensemble method \cite{Zhou2012}. Besides random sampling, it is interesting to explore smarter sampling strategies, such as BLB \cite{Kleiner:Talwalkar:Sarkar2012}, or feature hashing \cite{Weinberger:Dasgupta:Langford:Smola:Attenberg2009} when adequate. The \textit{hard negative mining} strategy may help improve generalization performance, and the effort improving the efficiency of hard negative mining may also be found helpful for the multi-grained scanning process \cite{Henriques:Carreira:Caseiro2013}. The efficiency of gcForest may be further improved by reusing some components during the process of different grained scanning, class vectors generation, forests training, completely-random trees generation, etc. In case the learned model is big, it may be possible to reduce to a smaller one by using the twice-learning strategy  \cite{Zhou:Jiang2004}; this might be helpful not only to reduce storage but also to improve prediction efficiency.

The employment of completely-random tree forests not only helps enhance diversity, but also provides an opportunity to exploit unlabeled data. Note that the growth of completely-random trees does not require labels, whereas label information is only needed for annotating leaf nodes. Intuitively, for each leaf node it might be able to require only one labeled example if the node is to be annotated according to the majority cluster on the node, or one labeled example per cluster if all clusters in the node are innegligible. This also offers gcForest with the opportunity of incorporating active learning \cite{Freund:Seung:Shamire:Tishby1997,Huang:Jin:Zhou2014} and/or semi-supervised learning strategies \cite{Zhou:Li2010,Li:Zhou2007}.

\section{Conclusion}\label{sec:conclusion}\vspace{-4mm}

In this paper, we try to address the question that \textit{Can deep learning be realized with non-differentiable modules}? We conjecture that behind the mystery of deep neural networks there are three crucial characteristics, i.e., layer-by-layer processing, in-model feature transformation, and sufficient model complexity, and we try to endow these characteristics to a non-NN style deep model. We propose the gcForest method\footnote{A shared code of gcForest is available at http://lamda.nju.edu.cn/code\_gcForest.ashx.} which is able to construct \textit{deep forest}, a deep model based on decision trees, and the training process does not rely on backpropagation. Comparing with deep neural networks, the gcForest has much fewer hyper-parameters, and in our experiments excellent performance is obtained across various domains by using even the same parameter setting. Note that there are other possibilities to construct deep forest. As a seminal study, we have only explored a little in this direction. Indeed, the most important value of this paper lies in the fact that it may open a door for non-NN style deep learning, or deep models based on non-differentiable modules.

In experiments we find that gcForest is able to achieve performance highly competitive to deep neural networks on a broad range of tasks. On some image task, however, its performance is inferior. On one hand, we believe that the performance of gcForest can be significantly improved, e.g., by designing better feature re-representation scheme rather than using the current simple classification vectors. On the other hand, it should not be ignored that deep neural network models such as CNNs have been investigated for more than twenty years by huge crowd of researchers/engineers whereas deep forest is just born. Furthermore, image tasks are killer applications of DNNs. It is generally too ambitious to aim at beating powerful techniques on their killer applications; e.g., linear kernel SVMs are still state-of-the-art for text categorization although DNNs have been hot for many years. Indeed, deep forest is not developed to replace deep neural networks; instead, it offers an alternative when deep neural networks are not superior, e.g., when DNNs are inferior to random forest and XGBoost. There are plenty of tasks where deep forests can be found useful.

\bibliographystyle{plain}
\bibliography{ensemble,deep}

\end{document}